\documentclass[letterpaper, 10 pt, conference]{ieeeconf}  % Comment this line out if you need a4paper
\usepackage{multirow}   %导言区
\usepackage{array}%需要用到该宏包
\usepackage{footnote}
\usepackage{amsmath}
\usepackage{amssymb}
\usepackage{graphicx}
\usepackage{cite}
\usepackage{booktabs}
\usepackage{tabularx}
\usepackage{supertabular}
\usepackage{color}
\makeatletter
\let\NAT@parse\undefined
\makeatother
\usepackage[colorlinks=true,
citecolor=blue,
linkcolor=blue,
anchorcolor=blue,
urlcolor=blue]{hyperref}
\IEEEoverridecommandlockouts                              % This command is only needed if 
\title{\LARGE \bf
MSDNet: Efficient 4D Radar Super-Resolution via Multi-Stage Distillation
}

\author{Minqing Huang\textsuperscript{*}, Shouyi Lu\textsuperscript{*}, Boyuan Zheng, Ziyao Li, Xiao Tang, Guirong Zhuo$^\dagger$ 
\thanks{This work was supported by the National Natural Science Foundation of China under Grant 52325212. \textsuperscript{*}The two authors contributed equally to this work. $\dagger$ indicates the corresponding author: Guirong Zhuo (zhuoguirong@tongji.edu.cn). The Authors are with School of Automotive Studies, Tongji University, Shanghai 201804, China. }
}
\begin{document}

\maketitle
\thispagestyle{empty}
\pagestyle{empty}

\begin{abstract}
%background
4D radar super-resolution, which aims to reconstruct sparse and noisy point clouds into dense and geometrically consistent representations, is a foundational problem in autonomous perception. However, existing methods often suffer from high training cost or rely on complex diffusion-based sampling, resulting in high inference latency and poor generalization, making it difficult to balance accuracy and efficiency. 
% contribution
To address these limitations, we propose MSDNet, a multi-stage distillation framework that efficiently transfers dense LiDAR priors to 4D radar features to achieve both high reconstruction quality and computational efficiency. The first stage performs reconstruction-guided feature distillation (RGFD), aligning and densifying the student’s features through feature reconstruction. In the second stage, we propose diffusion-guided feature distillation (DGFD), which treats the stage-one distilled features as a noisy version of the teacher's representations and refines them via a lightweight diffusion network. Furthermore, we introduce a noise adapter that adaptively aligns the noise level of the feature with a predefined diffusion timestep, enabling a more precise denoising.
% experiments
Extensive experiments on the VoD and in-house datasets demonstrate that MSDNet achieves both high-fidelity reconstruction and low-latency inference in the task of 4D radar point cloud super-resolution, and consistently improves performance on downstream tasks. The code will be publicly available upon publication.
\end{abstract}
\section{INTRODUCTION}
\label{sec:intro}
4D millimeter-wave (mmWave) radar, hereafter referred to as \emph{4D radar}, provides 3D point clouds together with instantaneous velocity measurements. It has become a key sensor in autonomous driving, and has attracted substantial attention in both academia and industry \cite{lu2025dnoi,huang2024low,peng2024transloc4d,ding2024milliflow,ding2024radarocc,zhang2025dual}. Due to its longer operating wavelength, 4D radar is largely insensitive to fine particulates (e.g., fog, drizzle), enabling stable perception in adverse weather conditions. Such robustness makes it well suited for all-weather autonomous driving.

\begin{figure}[t]
	\centering
	\includegraphics[width=0.45\textwidth]{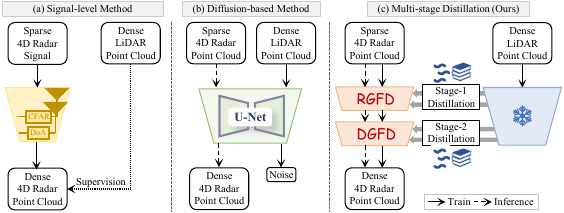}
	               \vspace{-2mm}
    \caption{Comparison of 4D Radar Super-Resolution Paradigms. (a) Signal-level methods replace classical signal-processing pipelines with deep neural networks. (b) Diffusion-based methods are trained on LiDAR point clouds and, during inference, iteratively denoise sparse 4D radar point clouds to produce dense point clouds. (c) Our MSDNet introduces a multi-stage distillation strategy that transfers LiDAR geometric priors to 4D radar and generates dense point clouds.}
	\label{fig0}
\vspace{-2mm}
\end{figure}

Despite these advantages, 4D radar data remain spatially sparse, noisy, and non-uniformly sampled compared to high-resolution LiDAR point clouds, severely limiting their effectiveness in fine-grained perception tasks such as object detection, scene understanding, and localization\cite{fan20244d}. To address these issues, as shown in Fig.\ref{fig0}(a), several studies process 4D radar signals directly, replacing traditional direction-of-arrival (DoA) estimation\cite{jiang20234d,zhao2022doa} and constant false-alarm rate (CFAR) pipelines with deep neural networks to enhance point cloud quality\cite{franceschi2022deep,han2024denserradar}; however, these approaches typically incur substantial data collection costs and exhibit limited transferability.

In recent years, diffusion models have achieved notable success in generative AI tasks\cite{cazenavette2024fakeinversion,li2024sd4match}. Building on these advances, recent work has further explored diffusion-based super-resolution for 4D radar point clouds. As illustrated in Fig.\ref{fig0}(b), these methods \cite{luan2024diffusion,zheng2025r2ldm} typically employ a U-Net backbone within the diffusion process. During training, these methods apply forward diffusion to LiDAR point clouds to train a noise prediction model. At inference, dense point clouds are generated from sparse 4D radar inputs via iterative reverse denoising. However, the significant distribution and density gap between 4D radar and LiDAR often requires complex diffusion architectures and multi-step sampling schedules to achieve competitive reconstruction quality, resulting in high inference latency.

In this paper, we propose \emph{Multi-Stage Distillation Net} (MSDNet), a knowledge distillation framework for 4D radar super-resolution. As illustrated in Fig.\ref{fig0}(c), MSDNet transfers knowledge from high-resolution LiDAR to 4D radar representations in two progressive stages. \textbf{Stage~1} employs reconstruction-guided feature distillation to transfer dense LiDAR features through feature reconstruction. \textbf{Stage~2} treats the Stage-1 distilled features as a ``noisy version" of the dense LiDAR features and applies a lightweight diffusion model equipped with prior modeling and iterative denoising to refine student representations. To properly initialize the diffusion process, we introduce a noise adapter module that dynamically estimates the noise level for each student feature and injects the corresponding Gaussian noise to match the appropriate diffusion timestep. Benefiting from the knowledge transferred in Stage~1, Stage~2 requires only a lightweight diffusion network, thereby significantly improving super-resolution quality while maintaining inference efficiency.

Overall, the contributions of our paper are as follows:
\begin{itemize}

\item We introduce \emph{Multi-Stage Distillation Net} (MSDNet), a knowledge-distillation framework for 4D radar super-resolution that efficiently transfers knowledge from dense LiDAR features to sparse 4D radar representations. To our knowledge, this is the first application of knowledge distillation to 4D radar super-resolution.
\item We design reconstruction-guided feature distillation for the initial transfer of dense LiDAR features, and introduce diffusion-guided feature distillation to refine the first-stage distilled features. In addition, we incorporate a noise adapter to precisely align the student features’ noise level with the initial timestep.
\item We conduct a comprehensive evaluation of MSDNet on the VoD \cite{palffy2022multi} and in-house datasets. Results show that MSDNet outperforms existing methods across reconstruction metrics while maintaining efficient inference. Moreover, MSDNet-reconstructed point clouds deliver significant gains on downstream tasks.

\end{itemize}
\section{RELATED WORK}
\label{sec:related}
\subsection{Knowledge Distillation for 4D Radar}

Knowledge distillation (KD)\cite{hinton2015distilling} provides a unified paradigm for cross-modal and cross-cost sensor transfer, enabling dense geometric and structural priors from LiDAR and cameras to be injected into 4D radar. Recent LiDAR-to-4D-Radar KD methods\cite{bang2024radardistill,xu2025sckd,song2025novel,wang2024self} have achieved notable gains in object detection, strengthening the discriminative power of sparse 4D radar point clouds. However, these frameworks are typically tightly coupled to detection heads and task-specific losses, directing optimization toward detection accuracy rather than point cloud resolution. Consequently, they are ill-suited for direct application to 4D point cloud super-resolution. Motivated by this gap, we adopt a reconstruction-centric perspective and develop a multi-stage distillation framework that progressively aligns and constrains representations across feature hierarchies to systematically improve 4D radar point cloud quality.

\subsection{4D Radar Point Cloud Super-Resolution}

Research on 4D radar point cloud super-resolution generally falls into two categories: (i) \emph{signal-chain} super-resolution, which replaces or augments traditional array processing (e.g., DoA, CFAR) during signal formation; and (ii) \emph{post-processing} super-resolution, which directly enhances the resolution and spatial regularity of 4D radar point clouds. On the signal-chain side, Zhao et al.\cite{zhao2022doa} perform end-to-end angular-occupancy regression from the upper-triangular portion of the array-output covariance matrix. Jiang et al.\cite{jiang20234d} achieve single-frame super-resolution DoA by combining a dual pulse-repetition-frequency waveform with a complex-valued convolutional network within a time-division– and Doppler-division–multiplexed MIMO framework. Denserradar trains directly on raw 4D radar tensors using dense occupancies synthesized from multi-frame LiDAR for supervision, while Franceschi et al. \cite{franceschi2022deep} replace the conventional CFAR front end with a CNN-based alternative. Despite improving 4D radar imaging, these approaches generally suffer from data scarcity and high training costs. On the post-processing side, building on advances in generative modeling, diffusion methods\cite{ho2020denoising,song2021denoising} have also been adopted for 4D radar super-resolution. Luan et al.\cite{luan2024diffusion} project 4D radar point clouds into BEV images and enhance them via BEV-based reconstruction. R2LDM\cite{zheng2025r2ldm} introduces a voxel-feature diffusion model that further improves point cloud detail. However, diffusion-based methods typically rely on U-Net backbones\cite{ronneberger2015u} and multi-step sampling, resulting in high inference latency. To overcome these bottlenecks, we propose a knowledge-distillation-based post-processing super-resolution framework for 4D radar. Operating under a teacher–student paradigm, it performs multi-stage feature distillation and employs a lightweight diffusion network to streamline the inference pipeline, substantially reducing latency and computational overhead while preserving reconstruction fidelity.

\begin{figure*}[t]
	\centering\includegraphics[width=0.99\linewidth]{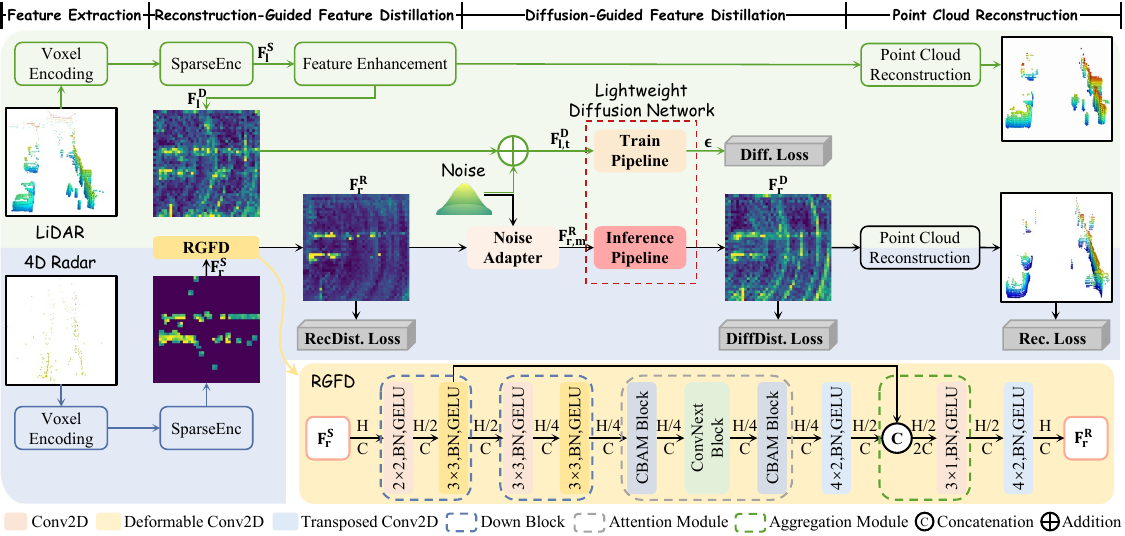} 
	\caption{Overall architecture of MSDNet. LiDAR and 4D radar point clouds are encoded by Voxel Encoding and SparseEnc to obtain BEV features, $\mathbf F_l^S$ and $\mathbf F_r^S$. 
    The teacher branch applies feature enhancement to produce dense LiDAR features $\mathbf F_l^D$ and to reconstruct the LiDAR point cloud. 
    The student branch first performs feature reconstruction on $\mathbf F_r^S$ within the RGFD module and conducts reconstruction distillation. Then, the DGFD module uses a lightweight diffusion model to refine the stage-one distilled features, producing the denoised representation $\mathbf F_r^D$. This representation is used for diffusion distillation and to reconstruct a dense 4D radar point cloud.
    The bottom right corner details the architecture of the RGFD module.}
\label{fig1}
\end{figure*}

 \section{PROPOSED METHOD}
\label{sec:method}
This section details MSDNet, with its overall architecture illustrated in Fig.~\ref{fig1}. Section~\ref{sec:3_1} provides a brief overview of the feature extraction and point cloud reconstruction networks employed as both teacher and student models. Section~\ref{sec:3_2} describes the construction of the teacher network. Section~\ref{sec:3_3} introduces the reconstruction-guided feature distillation (RGFD), which carries out the first-stage knowledge transfer through feature reconstruction. Finally, Section~\ref{sec:3_4} presents the diffusion-guided feature distillation (DGFD), which leverages a lightweight diffusion denoiser to further refine the student's features.
% \vspace{-2mm}
\subsection{Preliminary} 
\label{sec:3_1}
\textbf{Feature Encoding.} We employ VoxelNet \cite{zhou2018voxelnet} as a common feature encoder for both the LiDAR and 4D radar point clouds. Initially, the two types of point clouds are voxelized into a set of 3D voxels, $\mathbf V_\mathrm{mod}^\mathrm{3D}$. Subsequently, each branch is processed by a dedicated Voxel Feature Encoding and a sparse 3D convolutional backbone (collectively, $\text{SparseEnc}(\cdot)$). The resulting features are then aggregated along the height dimension to generate BEV representations, yielding two BEV features: $\mathbf F_\mathrm{l}^\mathrm{S}$ and $\mathbf F_\mathrm{r}^\mathrm{S}$:
\begin{equation}
	\label{eq_1}
	\mathbf F_\mathrm{mod}^\mathrm{S} = \Pi_z\left(\text{SparseEnc}\left(\mathbf V_\mathrm{mod}^\mathrm{3D}\right)\right), \mathrm{mod}\in\{\mathrm{l},\mathrm{r}\} ,
\end{equation}
where $\Pi_z(\cdot)$ denotes the aggregation along the height dimension, the subscripts $\mathrm{l}$ and $\mathrm{r}$ refer to the LiDAR and 4D radar branches. $\mathbf F_\mathrm{l}^\mathrm{D}$ and $\mathbf F_\mathrm{r}^\mathrm{D}$ are then obtained via feature enhancement and multi-stage feature distillation, respectively. 

\textbf{Point Cloud Reconstruction.} We adopt a progressive point cloud reconstruction module \cite{zheng2025r2ldm} to map dense BEV features to a 3D point cloud. Given the modal features $\mathbf F_\mathrm{mod}^\mathrm{D}$, we apply 3D upsampling layers to generate multi-scale voxel features at scales ${s}\in\{1/4,1/2,1\}$:
\begin{equation}
	\label{eq_2}
	\mathbf G_\mathrm{mod}^{(s)} = \text{Up3D}_{s}(\mathbf F_\mathrm{mod}^\mathrm{D}).
\end{equation}
At each scale, a dual-branch prediction head is employed to output the voxel occupancy and point offset, respectively:
\begin{equation}
	\label{eq_3}
	\mathbf M^{(s)} = \sigma\left(\mathrm\phi_\mathrm{mask}^{(s)}\left(\mathbf G_\mathrm{mod}^{(s)}\right)\right)\in \left[0,1\right].
\end{equation}
The offset branch predicts the point displacement relative to the center of its corresponding voxel:
\begin{equation}
	\label{eq_4}
	\Delta \mathbf P^{(s)} \!=\! \tanh\!\left(\phi_\mathrm{off}^{(s)}\left(\!\mathbf G_\mathrm{mod}^{(s)}\right)\!\right)\!\cdot\!\frac{\mathrm{L}^{(s)}}{2}\!\in\!\left(\!-\frac{\mathrm{L}^{(s)}}{2}\!,\!\frac{\mathrm{L}^{(s)}}{2}\!\right)^3,
\end{equation}
where $\sigma(\cdot)$ and $\tanh(\cdot)$ are the Sigmoid and Tanh activation functions, respectively. $\phi_\mathrm{mask}^{(s)}(\cdot)$ and $\phi_\mathrm{off}^{(s)}(\cdot)$ are the 3D convolutional branches for the occupancy and offset predictions. $\mathrm{L}^{(s)}$ is the side length of voxels at scale $s$. During inference, the point cloud is generated only at the full resolution $(s=1)$. Activated voxels are identified by applying a threshold to the occupancy map $\mathbf M^{(1)}$. The final point coordinates $\mathbf P^{(1)}$ are then computed by adding the predicted offsets $\Delta \mathbf P^{(1)}$ to the centers of these activated voxels $\mathbf C^{(1)}$:$\mathbf P^{(1)} = \mathbf C^{(1)} + \Delta \mathbf P^{(1)}$.

Notably, in our KD framework, the LiDAR branch serves as the teacher model, while the 4D radar branch acts as the student. The two branches employ distinct weights for feature extraction but share a common set of weights within the point cloud reconstruction network.
\subsection{Teacher Network}
\label{sec:3_2}
\textbf{Feature Enhancement.} Reconstructing point clouds directly from the sparse voxel features encoded by VoxelNet \cite{zhou2018voxelnet} struggles to recover fine-grained details. To address this, we introduce the S2D module from Sparse2Dense\cite{wang2022sparse2dense} on the teacher side, which acts as a LiDAR feature enhancer. This module refines the LiDAR features, $\mathbf F_\mathrm{l}^\mathrm{S}$, yielding dense teacher features, $\mathbf F_\mathrm{l}^\mathrm{D}$, that are richer in geometric and semantic information. 

\textbf{Teacher Loss.} The teacher network is trained with the objective of high-fidelity point cloud reconstruction. Given LiDAR point clouds as input, the network performs feature encoding and enhancement, followed by point cloud reconstruction, to generate multi-scale point clouds. The supervision signal comprises two terms—voxel occupancy and offset.

Let $N$ denote the total number of voxels. For the $i$-th voxel, let $\mathbf{M}_i\in[0,1]$ denote the predicted occupancy and ${\hat{\mathbf{M}}_i\in\{0,1\}}$ the ground-truth occupancy. The occupancy loss $\mathrm{\mathcal{L}_{\mathrm{occ}}}$ is defined as:   
\begin{equation}
\label{eq_6}
\begin{aligned}
\mathcal{L}_{\mathrm{occ}}
\!=\! -\frac{1}{N} \sum_{i=1}^{N} \!\biggl(\!
    \hat{\mathbf{M}}_{i}\log\!\left(\mathbf{M}_{i}\right)
    \!+\!\!\left(1\!\! -\!\! \hat{\mathbf{M}}_{i}\right)\!\log\!\left(1 \!\!- \!\!\mathbf{M}_{i}\right)\!
\biggr).
\end{aligned}
\end{equation}

The offset loss $\mathcal{L}_\mathrm{off}$ is calculated only over the $N_\mathrm{o}=\sum_i\hat{\mathbf{M}}_i$ occupied voxels. For each occupied voxel $i$ with center $\mathbf{C}_i\in \mathbb{R}^3$, predicted displacement $\Delta \mathbf{P}_i\in \mathbb{R}^3$, and ground-truth point coordinate $\hat{\mathbf{P}_i}\in \mathbb{R}^3 $, $\mathcal{L}_\mathrm{off}$ is defined as:
\begin{equation}
	\label{eq_7}
	\mathcal{L}_\mathrm{off}=\frac{1}{N_\mathrm{o}}\sum_{i=1}^{N_\mathrm{o}}\left\|\left(\Delta \mathbf{P}_i+\mathbf{C}_i\right)-\hat{\mathbf{P}}_i\right\|_1,
\end{equation}
where $\left\|\cdot\right\|_1$ denotes the L1 norm.

The total loss, $\mathcal{L}_\mathrm{teacher}$, is calculated as a weighted sum over $S=3$ scales:
\begin{equation}
	\label{eq_5}
	\mathcal{L}_\mathrm{teacher} = \sum_{s=1}^S\rho^{(s)}\mathcal{L}_\mathrm{occ}^{(s)}+\sum_{s=1}^{S}\zeta^{(s)}\mathcal{L}_\mathrm{off}^{(s)},
\end{equation}
where $\rho^{(s)}$ and $\zeta^{(s)}$ are the weights for the occupancy loss and the offset loss at scale $s$, respectively.

\subsection{Reconstruction-Guided Feature Distillation}
\label{sec:3_3}
The objective of Reconstruction-Guided Feature Distillation (RGFD) is to effectively transfer geometric priors from LiDAR to 4D radar through feature reconstruction. We observe that after VoxelNet \cite{zhou2018voxelnet} encoding, the number of non-empty grid cells in the 4D radar's BEV features is only about 10\% of that in the dense LiDAR features. This significant density disparity makes cross-modal alignment challenging. To this end, we propose a feature-reconstruction network that converts sparse 4D radar features into dense representations, thereby facilitating first-stage knowledge transfer.

\textbf{Network Architecture.} As shown in Fig.~\ref{fig1}, the RGFD network consists of two Down Blocks, two Up Blocks, an Attention Module, and an Aggregation Module. The Down Block combines standard and deformable convolutions to perform downsampling. This design enlarges the receptive field and alleviates feature misalignment induced by local deformations, while preserving high-response regions and filling gaps in the BEV grid. The Up Block progressively restores feature resolution using 2D transposed convolutions. The Attention Module consists of two CBAM modules\cite{woo2018cbam} that apply both channel and spatial attention to enhance salient features and suppress noise.  A ConvNeXt residual block\cite{liu2022convnet} is inserted between these two modules to further improve semantic aggregation and boundary refinement. The Aggregation Module concatenates two input features and applies a $3\times 1$ convolution layer. After applying the Down Block, Attention Module, and Up Block, in addition to incorporating a side pathway through the Aggregation Module, RGFD generates the reconstructed 4D radar features, $\mathbf{F}_\mathrm{r}^\mathrm{R}$.

\textbf{Reconstruction Distillation Loss.} To inject the dense geometric priors from the LiDAR into the 4D radar, the RGFD module is trained by minimizing the feature discrepancy between the reconstructed 4D radar features, $\mathbf{F}_\mathrm{r}^\mathrm{R}$, and the dense LiDAR features, $\mathbf{F}_\mathrm{l}^\mathrm{D}$. Let $\Omega_\mathrm{ne}$ and $\Omega_\mathrm{e}$ denote the sets of non-empty and empty grid cells in $\mathbf{F}_\mathrm{l}^\mathrm{D}$, with $N_\mathrm{ne}$ and $N_\mathrm{e}$ being the corresponding number of cells. The reconstruction distillation loss, $\mathcal{L}_\mathrm{distill}^\mathrm{rec}$, is defined as:
\begin{equation}
\label{eq_8}
\mathcal{L}^{\mathrm{rec}}_{\mathrm{distill}}
= \alpha\,\mathbb{E}_{i\in\Omega_\mathrm{ne}}\!\bigl[\|\Delta_\mathrm{R}^{\mathbf{F}(i)}\|_2\bigr]
+ \gamma\,\mathbb{E}_{i\in\Omega_\mathrm{e}}\!\bigl[\|\Delta_\mathrm{R}^{\mathbf{F}(i)}\|_2\bigr],
\end{equation}
where $\Delta_\mathrm{R}^{\mathbf{F}(i)}=\mathbf{F}^{\mathrm{D}}_{\mathrm{l}}(i)-\mathbf{F}^{\mathrm{R}}_\mathrm{r}(i)$, $\left\|\cdot\right\|_2$ denotes the L2 norm, $\alpha$ and $\gamma$ are the weights for the non-empty and empty feature terms, respectively.
\textbf{\begin{figure}[t]
	\centering
	\includegraphics[width=0.45\textwidth]{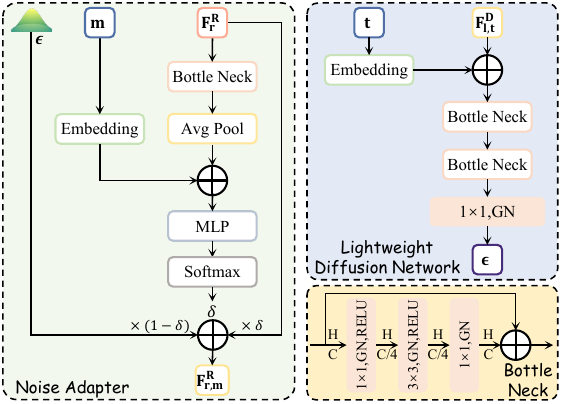}
	\caption{Details of the Noise Adapter and Lightweight Diffusion Network.}
	\label{fig2}
\end{figure}}
\subsection{Diffusion-Guided Feature Distillation}
\label{sec:3_4}
Although first-stage distillation substantially reduces cross-modal discrepancies, residual remain between student and teacher representations because of capacity differences \cite{huang2022knowledge}. We treat these residuals as noise in distilled features and therefore treat the reconstructed features $\mathbf{F}_\mathrm{r}^\mathrm{R}$ as a ``noisy version" of the dense LiDAR features $\mathbf{F}_\mathrm{l}^\mathrm{D}$.  
Building on this premise, we propose Diffusion-Guided Feature Distillation (DGFD), which learns a diffusion denoising prior from dense LiDAR features and applies it to the reconstructed features, producing a denoised representation $\mathbf{F}_\mathrm{r}^\mathrm{D}$ with finer-grained details.

\textbf{Teacher Side: Learning the Denoising Prior.} To learn the denoising prior, we subject the dense LiDAR features, $\mathbf{F}_\mathrm{l}^\mathrm{D}$, to a standard forward noising chain. The Markovian noising process is defined as:
\begin{equation}
	\label{eq_9}
	q(\mathbf{F}_{\mathrm{l},t}^\mathrm{D}|\mathbf{F}_{\mathrm{l}}^\mathrm{D}) = \mathcal{N}\left(\mathbf{F}_{\mathrm{l},t}^\mathrm{D} | \sqrt{\bar{\alpha}_t} \mathbf{F}_\mathrm{l}^\mathrm{D} ,\left( 1-\bar{\alpha}_t\right)\mathbf{I}\right),
\end{equation}
where $t\in \{0,1,\dots,T\}$, $\bar{\alpha}_t=\Pi_{s=1}^t\left(1-\beta_s\right)$,  $\{\beta_{s}\}_{\mathrm{s}=1}^T$ is a noise variance schedule,
 and $\mathbf{I}$ denotes the identity matrix. The diffusion network, $\Phi_\theta(\cdot,t)$, takes the noisy features $\mathbf{F}_{\mathrm{l},t}^D$ and the timestep $t$ as input to predict the noise term $\epsilon = \Phi_\theta(\mathbf{F}_{\mathrm{l},t}^\mathrm{D},t)$. The denoising prior is learned by minimizing the following L2 loss:
\begin{equation}
	\label{eq_10}
	 \mathcal{L}_\mathrm{diff} = \mathbb{E}_{t,\hat{\epsilon}_t}\left\| \hat{\epsilon}_t - \Phi_\theta\left( \mathbf{F}_{\mathrm{l},t}^\mathrm{D}, t \right) \right\|_2^2,
\end{equation}
where $\hat{\epsilon}_t\sim \mathcal{N}(0, \mathbf{I})$. This training process enables the diffusion network to accurately predict noise at each timestep, thereby establishing the foundation for the efficient denoising of the student's features.

\textbf{Student Side: Noise Adaptation and Denoising.} We treat the reconstructed features $\mathbf{F}_\mathrm{r}^\mathrm{R}$ as a ``noisy version" of the dense LiDAR features. However, the noise level within $\mathbf{F}_\mathrm{r}^\mathrm{R}$ is unknown and varies across samples, making it impossible to determine the initial timestep for the diffusion process directly. To resolve this, we propose a noise adapter module that aligns the noise level of the reconstructed features with that of a predefined timestep $m$. As shown in Fig.~\ref{fig2}, we first encode the timestep $m$ into an embedding feature. Concurrently, the reconstructed features $\mathbf{F}_\mathrm{r}^\mathrm{R}$ are passed through a bottleneck block and a Global Average Pooling (GAP) layer to extract a global semantic feature, $\mathbf{F}_\mathrm{r,g}^\mathrm{R}$:
\begin{equation}
	\label{eq_11}
	 \mathbf{F}_\mathrm{r,g}^\mathrm{R} = \mathrm{GAP}\left( \mathrm{BottleNeck}\left( \mathbf{F}_\mathrm{r}^\mathrm{R}\right) \right).
\end{equation}
Subsequently, the time embedding and the global semantic feature $\mathbf{F}_\mathrm{r,g}^\mathrm{R}$ are summed and fed into an MLP followed by a softmax to predict a gating coefficient, $\delta$. This coefficient is then used to linearly mix the reconstructed features with Gaussian noise $\hat{\epsilon} \sim \mathcal{N}(0,\mathbf{I})$, yielding a noisy initialization feature aligned with the predefined timestep $m$:
\begin{equation}
	\label{eq_12}
	 \mathbf{F}_{\mathrm{r},m}^\mathrm{R} = \delta \cdot \mathbf{F}_\mathrm{r}^\mathrm{R} + (1-\delta)\cdot \hat{\epsilon}.
\end{equation}
This feature is then fed into the diffusion network to perform iterative denoising:
    \begin{equation}
    \begin{aligned}
        p_\theta\left(\mathbf{F}_{\mathrm{r},t-n}^\mathrm{R}|\mathbf{F}_{\mathrm{r},t}^\mathrm{R}\right) = \mathcal{N}\left(\mathbf{F}_{\mathrm{r},t-n}^\mathrm{R} | \Phi_\theta\left(\mathbf{F}_{\mathrm{r},t}^\mathrm{R},t\right),\sigma_t^2\mathbf{I}\right),  
    \end{aligned}
        \label{eq_13}
        \end{equation}
where $t\in\{m,m-n,\dots,n\}$, $n$ represents the sampling interval. After the denoising process is complete, we obtain the denoised reconstructed features, $\mathbf{F}_\mathrm{r}^\mathrm{D}$. These are then used along with the dense LiDAR features, $\mathbf{F}_\mathrm{l}^\mathrm{D}$, to calculate the diffusion distillation loss, $\mathcal{L}_\mathrm{distill}^\mathrm{diff}$:
    \begin{equation}
        \!\mathcal{L}_\mathrm{distill}^\mathrm{diff} = \alpha\,\mathbb{E}_{i\in\Omega_\mathrm{ne}}\!\bigl[\|\Delta_\mathrm{D}^{\mathbf{F}(i)}\|_2\bigr]
+ \!\gamma\,\mathbb{E}_{i\in\Omega_\mathrm{e}}\!\bigl[\|\Delta_\mathrm{D}^{\mathbf{F}(i)}\|_2\bigr],  
    \label{eq_14}
    \end{equation}
where $\Delta_\mathrm{D}^{\mathbf{F}(i)}=\mathbf{F}^{\mathrm{D}}_{\mathrm{l}}(i)-\mathbf{F}^{\mathrm{D}}_\mathrm{r}(i)$.
%, $\left\|\cdot\right\|_2$ denotes the L2 norm, $\alpha$ and $\gamma$ are the weights for the non-empty and empty feature terms, respectively.

\textbf{Lightweight Diffusion Network.} The first-stage distillation has injected the geometric and structural priors of the dense LiDAR representations into the 4D radar representations. Therefore, in the second-stage distillation, we replace a heavyweight U-Net\cite{ronneberger2015u} with a lightweight diffusion network to accelerate inference without compromising noise regression accuracy. As shown in Fig.~\ref{fig2}, given a timestep $t$, we first encode it into a time embedding $\mathrm{e}_t$ and fuse it with the noisy features (either $\mathbf{F}_{\mathrm{l},t}^\mathrm{D}$ from the teacher side or $\mathbf{F}_{\mathrm{r},t}^\mathrm{R}$ from the student side). The fused features are then processed by two bottleneck blocks and an output head to predict the noise, $\epsilon$. Each bottleneck block preserves spatial resolution and transforms only the channel dimension: a $1\times1$ convolution compresses channels, a $3\times3$ convolution introduces local context, and a final $1\times1$ convolution restores the channel count, with a residual connection to stabilize training. This design achieves accurate noise regression with minimal parameters and computational cost.
\subsection{Loss Function}
\label{sec:3_5}
The denoised reconstructed features, $\mathbf{F}_\mathrm{r}^\mathrm{D}$, are fed into the point cloud reconstruction module to generate a dense 4D radar point cloud. Using the LiDAR point cloud as the ground truth, the student's reconstruction loss, $\mathcal{L}_\mathrm{recon}$, is then calculated following the same reconstruction paradigm as the teacher side.

Our KD framework jointly optimizes for both distillation and diffusion denoising. The diffusion branch learns a denoising prior via the diffusion loss, while the distillation branch learns a sparse-to-dense representation mapping through two distinct distillation losses. The total loss function for the student side is:
\begin{equation}
	\label{eq_15}
	  \mathcal{L}_\mathrm{student} \!=\! \lambda_1 \mathcal{L}_\mathrm{recon}\! +\! \lambda_2 \mathcal{L}_\mathrm{distill}^\mathrm{rec}\! + \!\lambda_3 \mathcal{L}_\mathrm{distill}^\mathrm{diff}\! + \!\lambda_4 \mathcal{L}_\mathrm{diff}.
\end{equation}
where $\lambda_1$, $\lambda_2$, $\lambda_3$ and $\lambda_4$ are the weights for each loss term.

\section{Implementation}
\label{sec:experiments}
\subsection{Dataset} 
We evaluate our method on the View-of-Delft (VoD) \cite{palffy2022multi} and an in-house dataset.

\textbf{VoD Dataset.} We follow 4DRVO-Net’s dataset partitioning scheme \cite{zhuo20234drvo} and adopt a more challenging setting to increase scene disparity between the training and test sets: sequences 03, 04, and 22 are held out for testing, and the remaining sequences are used for training.

\textbf{In-house Dataset.} To further assess generalization across scenes and 4D radar sensor, we construct an in-house dataset spanning nine closed-campus scenarios. The dataset comprises 20,720 synchronized frames captured by a 128-beam LiDAR and a 4D radar. We hold out sequence 09 (4,012 frames; two driving laps) for testing and use the remaining sequences for training.
\subsection{Implementation Details} 
For the BEV feature maps, the voxelization range is set to $[0,32]\times[-16,16]\times[-2,4]m$ along the $(x,y,z)$ axes. The voxel size is set to $(0.1\times0.1\times0.15)m$. We set the weights in Eq.~\ref{eq_5} to $\rho^{(1)}=\rho^{(2)}=\rho^{(3)}=1$ and  $\zeta^{(1)}=\zeta^{(2)}=\zeta^{(3)}=10$. In Eq.~\ref{eq_8} and Eq.~\ref{eq_14}, we use $\alpha=10$ and $\gamma = 20$. In Eq.~\ref{eq_15}, the weights are $\lambda_1=1$, $\lambda_2=0.01$, $\lambda_3=5$, $\lambda_4=10$. We leverage the DDIM\cite{song2021denoising} with a total of $T=1000$ diffusion steps. The predefined starting timestep $m=500$, sampling steps $T_m=50$. The sampling interval $n=10$ for Eq.~\ref{eq_13}. All experiments are conducted on a single NVIDIA RTX 4090 GPU. We use the Adam optimizer with an initial learning rate of $10^{-3}$ and a OneCycleLR scheduler. The batch size is set to 4. The teacher and student networks are trained for 60 and 90 epochs, respectively. Our data preprocessing follows R2LDM\cite{zheng2025r2ldm}, which includes removing ground points from the LiDAR data and cropping the LiDAR point cloud to match the Field of View (FOV) of the 4D radar.
\setlength{\tabcolsep}{1mm}
\begin{table*}[t]
\caption{Quantitative results on the VoD dataset \cite{palffy2022multi} and In-house dataset. The best result is \textbf{bold}, and the second best is \underline{underlined}.}	
        \vspace{-7mm}
	\centering
	\footnotesize
	\begin{center}
		\resizebox{1\textwidth}{!}
		{
			\begin{tabular}{l||ccccc||ccccc}
				\toprule
				&  \multicolumn{5}{c||}{VoD Dataset}  &\multicolumn{5}{c}{In-house Dataset} \\ 
				%\hline{2-25}
				\cline{2-11}\noalign{\smallskip}
				
				\multirow{-2}{*}{\begin{tabular}[c]{@{}c@{}}Method \end{tabular}}
				&  CD$\downarrow$  & MHD$\downarrow$ ($\times10^{-2}$)  & F-score$\uparrow$ & JSD$\downarrow$ & MMD$\downarrow$ ($\times10^{-4}$) &  CD$\downarrow$  & MHD$\downarrow$ ($\times10^{-2}$)  & F-score$\uparrow$ & JSD$\downarrow$ & MMD$\downarrow$ ($\times10^{-4}$)\\
				%\midrule
				%\cline{1-25}\noalign{\smallskip}
				\hline\hline
				\noalign{\smallskip}
				
				R2DM \cite{nakashima2024lidar} 
				&11.93 &276.28
				&0.12 &\underline{0.30}
				&\underline{11.63} &17.40
                &\underline{258.94} & 0.06
                & 0.50 & 45.29
				\\
                \hline 
				\noalign{\smallskip} 
				R2LDM \cite{zheng2025r2ldm}
				&\underline{6.07} &\underline{172.28}
				&\underline{0.27} &0.35
				&21.88 &\underline{9.67}
                &288.53 &\underline{0.13}
                &\underline{0.45} &\underline{19.49}
				\\
                \hline 
				\noalign{\smallskip} 
				MSDNet (Ours)
				&\bf{5.16}($\downarrow$15.0\%) 	&\bf{58.98}($\downarrow$65.8\%) 
				&\bf{0.39}($\uparrow$44.4\%)  	&\bf{0.21}($\downarrow$30.0\%)  
				&\bf{5.51}($\downarrow$52.6\%)   &\bf{6.59}($\downarrow$31.9\%)  
                &\bf{156.19}($\downarrow$39.7\%)   &\bf{0.17}($\uparrow$30.8\%)  
                &\bf{0.32}($\downarrow$28.9\%)   &\bf{7.49}($\downarrow$61.6\%)  
				\\ 
                \addlinespace[-\aboverulesep]
                \bottomrule
			\end{tabular}}
	\end{center}
	\label{table:sota}
\end{table*}

% \vspace{-1mm}
\subsection{Evaluation Metrics} 
We evaluate the model's performance from two aspects: 3D geometric accuracy and BEV spatial distribution consistency. For the former, we employ the Chamfer Distance (CD), Modified Hausdorff Distance (MHD)\cite{prabhakara2023high,zheng2025r2ldm}, and F-score\cite{tatarchenko2019single}. For the latter, we use the Jensen–Shannon Discrepancy (JSD) and Maximum Mean Discrepancy (MMD)\cite{zyrianov2022learning,nakashima2024lidar}. All evaluations use the preprocessed LiDAR point clouds as the reference ground truth.
% \vspace{-2mm}
\subsection{Baselines}
Given the limited availability of open-source methods for 4D radar point-cloud post-processing super-resolution, we select two representative baselines for comparison:

\textbf{R2DM}\cite{nakashima2024lidar} is a diffusion-based generative method for LiDAR point cloud super-resolution. In our study, we adapt this method to the 4D radar domain by using the 4D radar's range image as input.

\textbf{R2LDM}\cite{zheng2025r2ldm} is the current SOTA work for 4D radar point cloud super-resolution. It performs diffusion in a latent space of voxelized features to enhance the details of the reconstructed point cloud.
\begin{figure*}[t]
  \centering
  \includegraphics[width=0.95\linewidth, trim=0 0 0pt 0, clip]{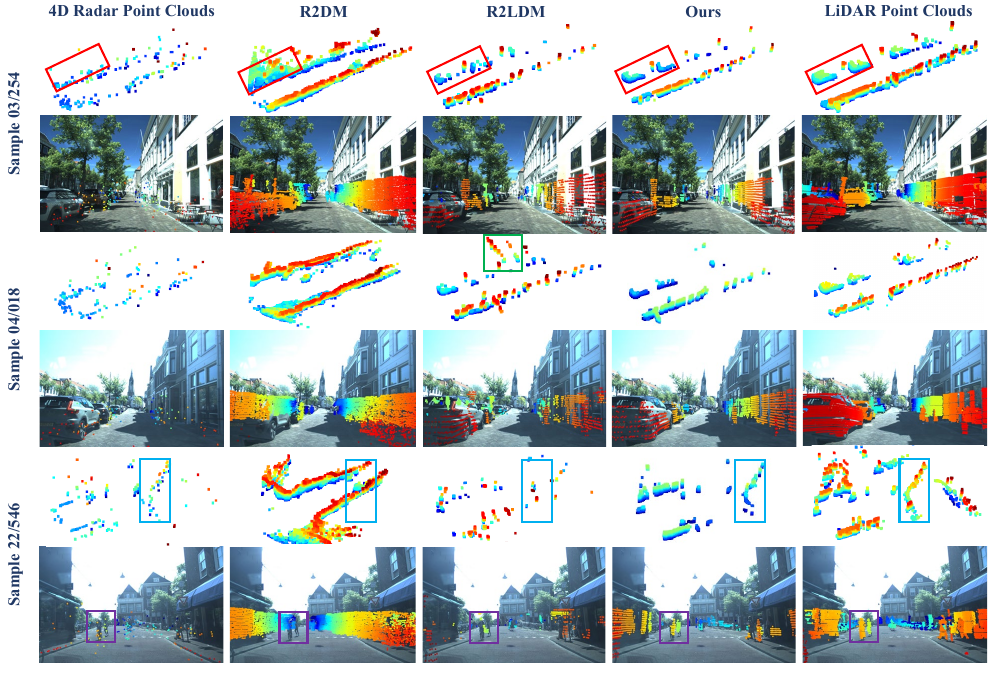}
  \caption{
     Qualitative results on the VoD dataset~\cite{palffy2022multi}.
  }
  \label{fig:results_vod}
\end{figure*}
\section{Experimental Results}
\subsection{Performance Evaluation} 
As shown in Tab.~\ref{table:sota}, MSDNet achieves SOTA results across all point cloud reconstruction metrics on both the VoD and in-house datasets. Both R2LDM and R2DM are diffusion-based super-resolution methods, but they are prone to generating hallucinatory points when under-constrained and overly reliant on training scene priors. Notably, R2DM, originally designed for LiDAR, is more sensitive to 4D radar noise, often resulting in denser point clouds but with inferior geometric fidelity. 

In contrast, MSDNet effectively transfers the geometric and structural priors from dense LiDAR via a two-stage distillation process. It robustly regularizes the 4D radar features through the dual constraints of feature reconstruction and diffusion denoising. This leads to significant improvements in the generated point cloud's global similarity, local fidelity, and BEV spatial distribution consistency. Specifically, on the VoD dataset, MSDNet achieves improvements of 65.8\% in MHD, 44.4\% in F-score, and 52.6\% in MMD. Our method also maintains leading performance on a more complex in-house dataset (with abundant trees and buildings), demonstrating excellent cross-scene and cross-sensor generalization.
\subsection{Ablation Study}
To analyze the effectiveness of each module in our proposed method, we conduct an ablation study on the VoD dataset by removing or replacing some key components. The results are presented in Tab.~\ref{table2} and Tab.~\ref{table3}.
\setlength{\tabcolsep}{0.9mm}
\begin{table}[t]
	\caption{Results of the ablation study.}
        \vspace{-2mm}
	\label{table2}
	\centering
	\footnotesize
	\resizebox{0.49\textwidth}{!}{
	\begin{tabular}{cc||cccc}
	\toprule
	\multicolumn{2}{c||}{Method} &CD$\downarrow$ & MHD$\downarrow$($\times10^{-2}$) & JSD$\downarrow$ & MMD$\downarrow$($\times10^{-4}$) \\ \hline\hline \noalign{\smallskip}
        \multirow{2}{*}{(a)} &w/o RGFD &5.88 &66.43 &0.22& 5.94 \\
	% \hline
        &w/o DGFD & 9.29 & 135.57 & 0.27 &12.03 \\
        \hline
        \multirow{2}{*}{(b)} & w/o Noise Adapter & 5.53 & 61.74 &0.22 & 6.21 \\
        & w/o Time Embedding &\underline{5.36} & \underline{60.40} &0.22 & 6.05 \\
        \hline
        (c) &with U-Net  & 6.13 & 101.88 & \bf{0.20} &\bf{5.48} \\
        \hline
        \multicolumn{2}{c||}{MSDNet (Full)} &\bf{5.16} &\bf{58.98} & \underline{0.21} &\underline{5.51} \\		
	\addlinespace[-\aboverulesep]
    \bottomrule
	\end{tabular}}
\end{table}
\setlength{\tabcolsep}{0.9mm}
\begin{table}[t]
	\caption{Ablation on diffusion sampling steps.}
    \vspace{-2mm}
	\label{table3}
	\centering
	\resizebox{0.49\textwidth}{!}{
		\begin{tabular}{c||cccc||c}
	\toprule
	$T_m$ &CD$\downarrow$ & MHD$\downarrow$($\times10^{-2}$) & JSD$\downarrow$ & MMD$\downarrow$($\times10^{-4}$) & Runtime(s) \\ \hline\hline \noalign{\smallskip} 
	3 &6.02	&73.26	&0.22	&6.48	&0.08s \\
		\hline
        10	&5.34	&65.94	&0.21	&5.63	&0.09s	\\
        \hline
        50	&\bf5.16	&\bf58.98	&\underline{0.21}	&\underline{5.51}	&0.10s \\
        \hline
        100	&\bf5.16	&\underline{64.07}	&\bf0.20	&\bf5.40	&0.13s \\
	\addlinespace[-\aboverulesep]
    \bottomrule
	\end{tabular}}
\end{table}
\setlength{\tabcolsep}{0.9mm}
\begin{table}[t]
       	\caption{Average runtime and parameters on the VoD dataset.}
    \vspace{-2mm}
	\label{table4}
	\centering
	\resizebox{0.38\textwidth}{!}{
		\begin{tabular}{c||c||c||c}
	\toprule
	Metrics &R2DM & R2LDM & MSDNet (Ours) \\ \hline\hline \noalign{\smallskip}
	Runtime(s)$\downarrow$ &0.76	&0.96	&0.10 \\
		\hline
        Parameters(M)$\downarrow$	&31.1	&40.3	&20.5	\\
	\addlinespace[-\aboverulesep]
    \bottomrule
	\end{tabular}}
\end{table}
\begin{figure}[t]
	\centering
	\resizebox{0.96\linewidth}{!}
	{

	\includegraphics[scale=0.80]{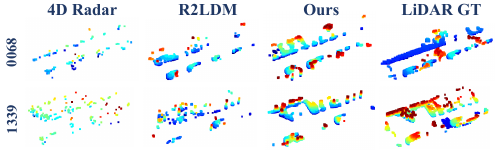}}
                % \vspace{-1mm}
\caption{Qualitative results on the in-house dataset.}
\label{fig:jihe}
% \vspace{-3mm}
\end{figure}
\setlength{\tabcolsep}{0.9mm}
\begin{table}[t]
	\renewcommand{\arraystretch}{1.2}
	\caption{4D Radar Odometry results on the VoD dataset.}
    \vspace{-2mm}
	\label{table5}
	\centering
	\resizebox{0.49\textwidth}{!}{
		\begin{tabular}{c||cc|cc|cc|cc}
			\toprule
			\multirow{2}{*}{Point Cloud Type} & \multicolumn{2}{c|}{03} & \multicolumn{2}{c|}{04} & \multicolumn{2}{c|}{22} & \multicolumn{2}{c}{Mean} \\ \cline{2-9} 
			& $t_{rel}$$\downarrow$          & $r_{rel}$$\downarrow$         & $t_{rel}$$\downarrow$          & $r_{rel}$$\downarrow$         & $t_{rel}$$\downarrow$          & $r_{rel}$$\downarrow$         & $t_{rel}$$\downarrow$          & $r_{rel}$$\downarrow$           \\ \hline\hline \noalign{\smallskip} 
	   4D radar & 0.43	&0.80	&0.19	&0.26	&0.28	&0.46	&0.30	&0.51 \\ \hline
       R2DM	& 0.79	& 1.74	& 0.84	& 0.73	& 0.65	& 1.49	& 0.76	& 1.32 \\ \hline
       R2LDM	& 0.56	& 1.05	& 0.55	& 0.41	& 0.57	& 0.52	& 0.56	& 0.66 \\ \hline
       MSDNet (Ours)	& \bf0.35	& \bf0.73	& \bf0.13	& \bf0.23	& \bf0.25	& \bf0.41	& \bf0.24	& \bf0.46 \\
			\addlinespace[-\aboverulesep]
            \bottomrule
	\end{tabular}}
% \vspace{-1mm}
\end{table}
\setlength{\tabcolsep}{0.9mm}
\begin{table}[!t]
	\renewcommand{\arraystretch}{1.2}
	\caption{Place recognition results on the in-house dataset.}
    \vspace{-2mm}
	\label{table6}
	\centering
	\resizebox{0.38\textwidth}{!}{
		\begin{tabular}{c||c|c|c}
			\toprule
			Point Cloud Type & Recall@1$\uparrow$ & Recall@3$\uparrow$ & Recall@5$\uparrow$\\ \hline\hline \noalign{\smallskip}
	   4D radar & 47.06	& 64.14	& 71.16 \\ \hline
       R2DM	& 11.20	& 21.26	& 26.29 \\ \hline
       R2LDM	& 50.89	& 65.13	& 70.80 \\ \hline
       MSDNet (Ours)	& \bf58.54	& \bf67.68	& \bf72.71 \\
			\addlinespace[-\aboverulesep]
            \bottomrule
	\end{tabular}}
% \vspace{-2mm}
\end{table}

\textbf{Benefits of the Multi-Stage Distillation Module.} As shown in Tab. \ref{table2}(a), RGFD transfers dense LiDAR feature priors to sparse 4D radar features via feature reconstruction, substantially enhancing their representational capacity. DGFD employs a diffusion model to remove residual noise from the distilled features, bringing them closer to dense LiDAR features. Consequently, all point cloud reconstruction metrics exhibit significant improvements.

\textbf{Benefits of Noise Adapter and Time Embedding.} As shown in Tab. \ref{table2}(b), with the introduction of noise adapter, the matching between the reconstructed features and the predefined noise level becomes more precise, leading to significant gains in reconstruction metrics. The time embedding module, by strengthening the correlation between features and their predefined timesteps, effectively boosts both noise matching accuracy and reconstruction fidelity.

\textbf{Different Diffusion Network.} As shown in Tab. \ref{table2}(c), we replaced our lightweight diffusion network with a U-Net\cite{ronneberger2015u} as the noise predictor. The average runtime reached 1.03s, which is significantly higher than the 0.10s achieved by our method; meanwhile, the performance on reconstruction metrics was comparable between the two. This result shows our lightweight design greatly cuts inference cost without losing reconstruction quality.

\textbf{Effect of Sample Steps.} As shown in Tab. \ref{table3}, we evaluate the impact of different numbers of sampling steps, $T_m$. $T_m=50$ achieves the best or second-best performance with a relatively small time overhead. Therefore, we adopt it as our default setting.
\subsection{Visualization}
Fig. \ref{fig:results_vod} and Fig. \ref{fig:jihe} present the qualitative comparisons on the VoD and in-house datasets, respectively. MSDNet is capable of reconstructing more complete and geometrically consistent structures from the sparse 4D radar point clouds, including road contours, side walls, vehicles, trees, pedestrians/cyclists, and distant small objects, with a more uniform point distribution. In contrast, the point clouds reconstructed by R2LDM suffer from an uneven distribution, an insufficient depiction of small objects, and a tendency to produce structural hallucinations (e.g., generating a non-existent turning lane). Although R2DM produces a higher point density, it lacks geometric accuracy and boundary fidelity, and is also prone to generating spurious structures.
\subsection{Runtime and Params} As shown in Tab. \ref{table4}, we compare the average inference time and parameters for each method on the VoD dataset. Owing to our two-stage distillation and lightweight diffusion model, MSDNet is 89.6\% faster than R2LDM and has only half the parameters. This indicates that our method not only improves point cloud reconstruction accuracy but also substantially reduces the model’s runtime and parameters.
\subsection{Performance on Downstream Task} 
\textbf{4D Radar Odometry.} Odometry accuracy is correlated with the clarity and geometric fidelity of the point cloud. We conduct an odometry evaluation on the VoD dataset using the GICP algorithm \cite{segal2009generalized} on both the raw 4D radar point clouds and the point clouds generated by the methods. The evaluation metrics are adopted from 4DRO-Net\cite{lu2023efficient}. As shown in Tab. \ref{table5}, the point clouds generated by MSDNet achieve the best results across all test sequences.

\textbf{4D Radar Place Reconition.} Place recognition performance is highly dependent on the richness and accuracy of the point cloud descriptors. We use ScanContext\cite{kim2018scan} for place recognition on our in-house dataset, and the evaluation protocol follows that of TDFANet\cite{lu2025tdfanet}. As shown in Tab. \ref{table6}, both MSDNet and R2LDM yield significant gains over the raw radar point cloud, with MSDNet comprehensively outperforming R2LDM.
\section{CONCLUSION}
In this paper, we presented MSDNet, a multi-stage distillation framework for 4D radar super-resolution that progressively transfers dense LiDAR priors to the 4D radar representation through reconstruction-guided and diffusion-guided stages. This dual-stage approach significantly reduces inference latency and parameter size while maintaining high reconstruction fidelity. Extensive experiments demonstrate the superiority and strong generalization of MSDNet, as well as its benefits on downstream tasks.

\bibliographystyle{IEEEtran}
\bibliography{ref}
\end{document}